\documentclass[journal]{IEEEtran}

\ifCLASSINFOpdf
\else
\fi
\usepackage{amsmath}
\usepackage{amssymb}
\usepackage{graphicx}
\usepackage{multirow}
\usepackage{booktabs}
\usepackage{wrapfig}
\usepackage{xspace}
\usepackage{hyperref}
\usepackage{balance}
\usepackage{multicol}
\usepackage{diagbox}
\usepackage{float}
\usepackage{ulem}
\usepackage{xcolor}
\usepackage{colortbl}
\newcommand{\blueline}[1]{{#1}}

\begin{document}


\title{RoboDuet: \blueline{Learning a Cooperative Policy for Whole-body Legged Loco-Manipulation}}

\author{Guoping Pan,
        Qingwei Ben,
        Zhecheng Yuan,
        Guangqi Jiang,
        Yandong Ji,\\
        Shoujie Li,
        Jiangmiao Pang,
        Houde Liu,~\IEEEmembership{Member,~IEEE,}
        Huazhe Xu
\thanks{This work was supported by Institute for Interdisciplinary Information Sciences, Tsinghua University, Shanghai Qi Zhi Institute and Shanghai Artificial Intelligence Laboratory.(\textit{Guoping Pan and Qingwei Ben are co-first authors.})(\textit{Corresponding author: Huazhe Xu.})}

\thanks{Guoping Pan, Qingwei Ben, Zhecheng Yuan and Shoujie Li are with Tsinghua University, Beijing 100084, China.(e\-mail:pgp23@mails.tsinghua.edu.cn, elgceben@gmail.com, yuanzc23@mails.tsinghua.edu.cn, lsj20@mails.tsinghua.edu.cn)}
\thanks{Guangqi Jiang and Yandong Ji are with University of San Diego, California 92093, USA.(e\-mail:gqjiang@ucsd.edu, ydji1024@gmail.com)}
\thanks{Jiangmiao Pang is with Shanghai Artificial Intelligence Laboratory, Shanghai 200032, China.(e\-mail:pangjiangmiao@gmail.com)}
\thanks{Houde Liu is with Tsinghua Shenzhen International Graduate School, Shenzhen 518055, China, and also with Jianghuai Advance Technology Center, Hefei 230000.(e\-mail:liu.hd@sz.tsinghua.edu.cn)}
\thanks{Huazhe Xu is with the Institute for Interdisciplinary Information Sciences, Tsinghua University, Beijing 100084, China, and also with Shanghai Qi Zhi Institute, Shanghai 200030, China, as well as with Shanghai Artificial Intelligence Laboratory, Shanghai 200032, China.(e\-mail: huazhe\_xu@mails.tsinghua.edu.cn)}
}

\newcommand{\ourshort}{RoboDuet\xspace}

\maketitle

\begin{abstract}
Fully leveraging the loco-manipulation capabilities of a quadruped robot equipped with a robotic arm is non-trivial, as it requires controlling all degrees of freedom (DoFs) of the quadruped robot to achieve effective whole-body coordination. In this letter, we propose a novel framework \ourshort, which employs two collaborative policies to realize locomotion and manipulation simultaneously, achieving whole-body control through mutual interactions. Beyond enabling large-range 6D pose tracking for manipulation, \blueline{we find that the two-policy framework supports zero-shot transfer across quadruped robots with similar morphology and physical dimensions in the real world.} Our experiments demonstrate that \ourshort achieves a \blueline{23\%} improvement in success rate over the baseline in \blueline{challenging} loco-manipulation tasks employing whole-body control. To support further research, we provide open-source code and additional videos on our website: \href{https://locomanip-duet.github.io/}{locomanip-duet.github.io}.

\end{abstract}

\IEEEpeerreviewmaketitle

\section{Introduction}




\IEEEPARstart{M}obile robots have increasingly been deployed to assist humans and demonstrated remarkable capabilities~\cite{fu2024mobile,wu2023tidybot}. Typically, these robots are equipped with wheeled mobile bases, making them less adaptable to diverse terrains and unable to adjust their base postures. This limitation has sparked interest in developing legged robots to undertake manipulation tasks, offering enhanced versatility and adaptability in diverse environments. By employing whole-body control in legged robots and robotic arms, it is possible to effectively overcome terrain constraints and significantly expand the manipulation workspace of the arms~\cite{fu2023deep,zhang2023gamma,qiu2024learning}. However, training a legged loco-manipulation policy to achieve whole-body control, along with accurate 6D pose tracking capabilities, presents a substantial challenge to researchers.

\begin{figure}
    \centering
    \includegraphics[width=0.48\textwidth]{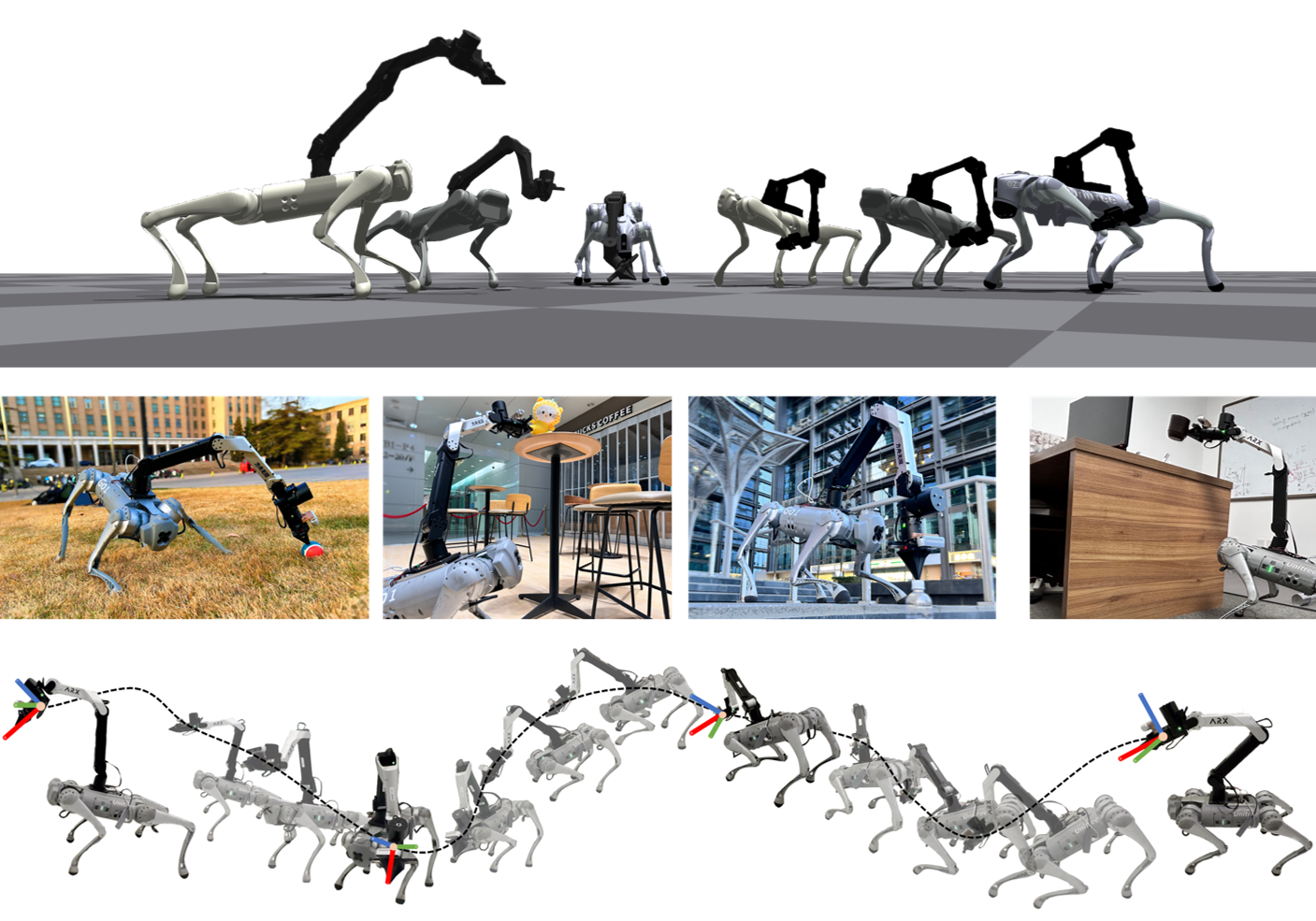}
    
    \caption{\ourshort is a framework that affords loco-manipulation and \blueline{zero-shot transfer on morphologically and dimensionally similar quadruped robots. Top row: zero-shot transfer.} This feature enables control across six configurations generated by the permutation of two robotic arms and three quadruped robots, allowing for \blueline{the replacement of quadruped robots} without the need to retrain the entire system. Middle row: loco-manipulation. From left to right, the robot walks to pick up a small ball on grass, grasps a doll from a high table in a café, grabs a bottle on lower stairs, and picks up a cup from an office desk. Bottom row: whole-body control. Given multiple target end-effector poses, the robot can adjust its entire body posture while moving to reach the desired targets and maintain stability.}
    \label{fig:big}
\end{figure}

As a pioneering effort in this domain, Fu \textit{et al}.~\cite{fu2023deep} has utilized a unified control policy to accomplish coordinated manipulation and locomotion. Despite the implementation of a whole-body control framework, it cannot tackle accurate 6D end-effector pose tracking, a capability that is crucial for manipulation tasks.
On the other hand, while GAMMA~\cite{zhang2023gamma} and GeFF~\cite{qiu2024learning} are capable of grasping objects based on 6-DoF end-effector control, their operation strategies separate the arm and the quadruped systems, thus falling short of achieving whole-body control. This distinction restricts the workspace of the arm. Consequently, accomplishing large-range manipulation tasks across the entire workspace necessitates a novel training paradigm. This approach must ensure enhanced coordination between the quadruped and the manipulator arm while also improving training efficiency and generalization capabilities. 



In awareness of these challenges, we introduce the \textbf{\ourshort}: an integrated legged loco-manipulation framework tailored for large-range 6D pose tracking and whole-body contorl. As shown in Fig. \ref{fig:big}, RoboDuet endows the policies with the capabilities of \blueline{zero-shot transfer across quadruped robots with similar morphology and physical dimensions}, loco-manipulation for diverse tasks, and robust whole-body control. These capabilities greatly enhance the system’s adaptability to different environments and reduce the need for retraining. 

\begin{figure*}
    \centering
    \vspace{0.2em}
    \includegraphics[width=0.9\textwidth]{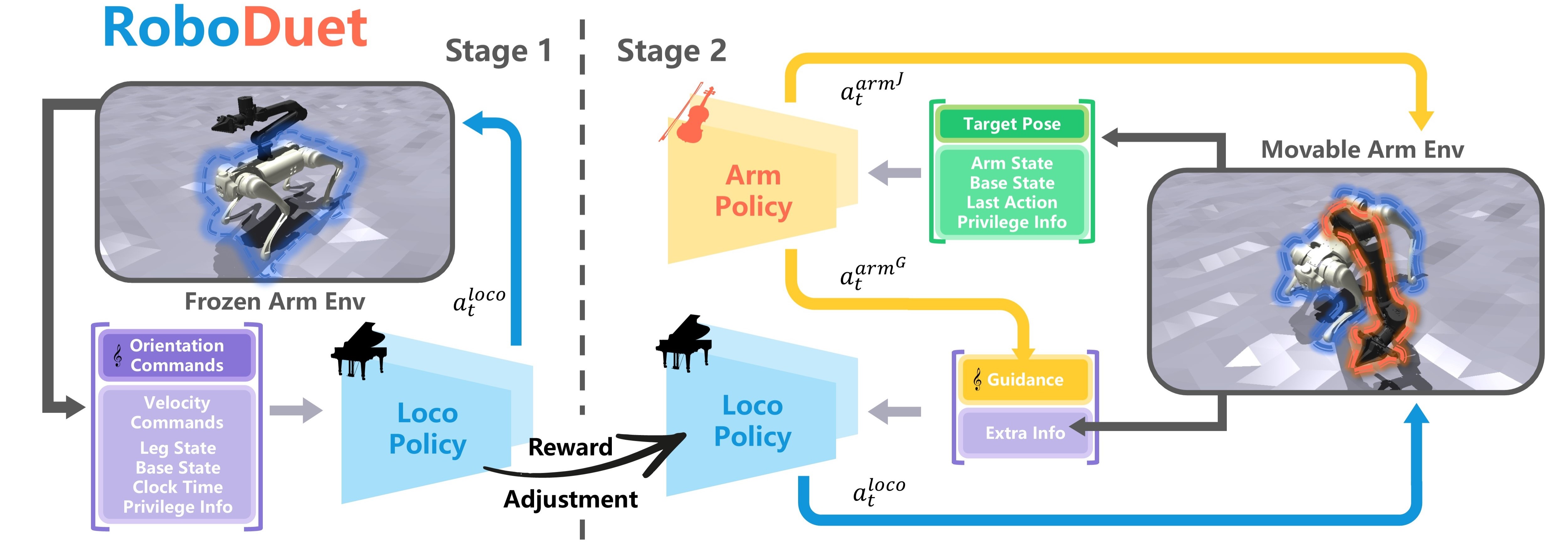}
    \caption{Overview of RoboDuet. In Stage 1, the loco policy is trained with fixed robotic arm, enabling the quadruped robot to achieve robust locomotion. In Stage 2, the loco and arm policies are trained simultaneously in a cooperative manner. The loco policy from stage 1 is reused, but the original body orientation commands are replaced by guidance signals $a_t^{arm^G}$ generated by the arm policy, enabling coordinated whole-body control. During the transition between the two stages, reward adjustment is introduced to effectively leverage locomotion priors, facilitating seamless integration for whole-body loco-manipulation.}
    \label{fig:framework}
\end{figure*}

To simplify the whole-body control problem, we adopt a cooperative mechanism, achieved through the coordinated collaboration of a locomotion policy and an arm policy. The interaction between the locomotion policy and the arm's actions exhibits a duet-like performance, where the locomotion policy utilizes the actions of the arm as guidance to adjust its posture, while the arm is complemented by the actions of the locomotion policy aiming to expand the robot's workspace.
The training process for \ourshort is structured in two stages, as illustrated in Fig. \ref{fig:framework}. In stage 1, we develop the locomotion policy to endow the legged robot with essential mobility capabilities. 
Following stage 1, stage 2 involves training the arm policy that can coordinate with the locomotion policy. We argue that employing a two-stage training strategy enhances the stability of the training process, resulting in the acquisition of agile and large-range 6D pose tracking agents.


Our contributions are summarized as follows:
\begin{itemize}
    \item \blueline{We propose a novel RL-based whole-body control framework for quadrupedal loco-manipulation that enables 6D end-effector pose tracking during locomotion.}
    \item \blueline{We introduce a cooperative mechanism that decouples multi-task learning, allowing the two policies to collaborate while maintaining focus and resulting in enhanced loco-manipulation capabilities compared to the unified policy.}
    \item \blueline{Our approach achieves at least a 23\% improvement in success rate for challenging mobile manipulation tasks over model-based controllers and demonstrates zero-shot transfer capabilities across quadruped robots with a 2.7 kg increase in weight.}
\end{itemize}

\section{Related Works}

\blueline{In recent years, the capabilities of legged robot locomotion have advanced rapidly, particularly in traversing complex terrains~\cite{grandia2023perceptive, zhuang2023robot}, thereby facilitating their integration into diverse and challenging application domains.} Furthermore, given the redundant degrees of freedom of quadruped robots compared to wheeled or tracked platforms, 
there are growing research efforts focused on whole-body control for legged robots with manipulators to achieve loco-manipulation capabilities. The current technological landscape features three primary strategies. \blueline{The first approach leverages control-based techniques such as model predictive control (MPC), which explicitly consider the system’s kinematic and dynamic properties, making them well-suited for scenarios involving dense contact forces. However, these methods often entail substantial engineering effort and exhibit inherent limitations in adaptability and robustness when applied to unknown environments~\cite{SleimanUMPC,10.1007/s10514-023-10146-0, sleiman2023versatile}.} 
The second strategy adopts learning algorithms to generate high-level commands for legged robot or robotic arm, which are then translated into low-level joint control instructions based on built-in controllers or inverse kinematics (IK)~\cite{zhang2023gamma,qiu2024learning,yokoyama2023adaptive,liu2024visual}. However, these approaches lack effective coordination between the legged platform and the arm or suffer from infeasible IK solutions, thus failing to maximize the potential for coordinating the pose of the legged robot to extend the operational workspace of the arm. The third approach leverages deep reinforcement learning (DRL) to realize whole-body control. \blueline{In earlier research, a pipeline was developed to integrate a model-based manipulator with an RL-based locomotion policy, demonstrating robustness against external force disturbances induced by the manipulator~\cite{ma2022combining}. However, the locomotion policy focused solely on resisting external disturbances, lacking the ability to achieve whole-body coordination.} Later study proposed training a unified policy to control the entire system for achieving whole-body control, but this approach were limited to achieving only end-effector position tracking~\cite{fu2023deep}. A follow-up study extends this approach to 6D task-space pose tracking~\cite{ha2024umionlegs}, but it is limited by the constraints of data collection. The most recent research leverages keypoint tracking to train a whole-body end-effector pose tracking policy, which requires an additional policy to provide locomotion capabilities~\cite{portela2024whole}. \blueline{This indicates that the method lacks the ability to perform simultaneous mobility and manipulation, which is crucial for executing dynamic tasks such as door opening.} Consequently, there is a clear need for innovative frameworks that can fully harness the locomotive advantages of quadruped robots while ensuring seamless coordination between the upper arm and the lower legged platform.

\section{Methods}

\subsection{Cooperative Policy for Whole-body Control}
\ourshort consists of a loco policy for locomotion and an arm policy for manipulation. The two policies are harmonized as a whole-body controller. Specifically, the loco policy adjusts its actions by following the instructions of the arm policy. 

\subsubsection{\textbf{Loco policy}} The goal of the loco policy $\pi_{loco}$ is to follow a target command $\mathbf{c}_t=(v^{\mathrm{cmd}}_x, v^{\mathrm{cmd}}_y, {\omega}^{\mathrm{cmd}}_{yaw}, \phi^{\mathrm{cmd}}_{pitch}, \phi^{\mathrm{cmd}}_{roll})$, where $v^{\mathrm{cmd}}_{x,y} = (v^{\mathrm{cmd}}_x, v^{\mathrm{cmd}}_y)$ is the desired linear velocity in base frame along x-  and y- axes, ${\omega}^{\mathrm{cmd}}_{yaw}$ is the desired angular velocity in yaw axis, $\boldsymbol{\phi}^{\mathrm{cmd}} = (\phi^{\mathrm{cmd}}_{pitch}, \phi^{\mathrm{cmd}}_{roll})$ denotes the desired pitch and roll angles of the base which plays an important role in achieving whole-body control and many downstream tasks.
The observation of loco policy $o_{t}^{loco}$ contains leg states \blueline{$s^{leg}_t \in \mathbb{R}^{24}$} (leg joint positions and velocities),  base states $\boldsymbol{\phi}_t$ (roll and pitch angles), target commands $\mathbf{c}_t$, clock time $\mathbf{t}_t$, and last leg action $a^{leg}_{t-1} \in \mathbb{R}^{12}$. The leg action $a_t^{leg}$ represents joint position offsets that are added to the default joint positions to specify the target positions for the twelve leg joint motors.

\subsubsection{\textbf{Arm policy}} The goal of the arm policy $\pi_{arm}$ is to accurately track the 6D end-effector pose. The observation of arm policy $o_{t}^{arm}$ is composed of arm states $s^{arm}_t \in \mathbb{R}^{12}$ (arm joint positions and velocities), target end-effector pose \blueline{$\chi_t \in \mathbb{R}^9$}, 
base states $\boldsymbol{\phi}_t$, and last arm action $a^{arm}_{t-1} \in \mathbb{R}^{8}$. 
The actions of the arm policy consist of two parts: the first six actions $a^{arm^J}_t \in \mathbb{R}^6$ represent the target joint position offsets corresponding to six arm joint actuators. Then, it will be concatenated with the output of the loco policy to achieve synchronous control of the overall system. Notably, the position targets are tracked using a proportional-derivative controller. To expand the manipulation workspace with whole-body control, the rest part of the arm policy $a_t^{arm^G}=(a_t^{arm^G_p}, a_t^{arm^G_r}) \in \mathbb{R}^2$ is used to replace $\boldsymbol{\phi}^{\mathrm{\mathrm{cmd}}}$,
providing additional degrees of freedom for pose tracking to cooperate with the loco policy.
Due to the distinct focuses of the two policies, only essential information is shared to maintain a favorable cooperative relationship while mitigating disruptions caused by factors such as differences in task learning efficiency.

\subsection{Two-stage Training}

To ensure robust loco-manipulation, we employ a two-stage training strategy. Our method enables a seamless transition between these stages by maintaining consistent input and output dimensions for the policies throughout the entire process. Since the primary tasks differ across the stages—stage 1 focuses on locomotion, while stage 2 emphasizes manipulation—we introduce reward adjustment to integrate the latter without compromising the performance of the former.

\subsubsection{\textbf{Stage 1}} \label{subsubsec: stage-1} This stage focuses on obtaining the robust locomotion capability. To ensure that the loco policy adapt to the center of mass and the inertia offset of the whole robot throughout the entire training process, we keep all the arm joints fixed at their default positions $[0, 0.8, 0.8, 0, 0, 0]$. In this stage, the arm policy is inactive, and the target end-effector pose $\chi_t$ is set to zero. Inspired by the powerful blind locomotion algorithm \cite{margolis2023walk},  we similarly apply a vector of behavior parameters $\boldsymbol{\mathrm{b}}_t$ to represent a heuristic gait reward, which is expressed as follows: 
\begin{equation}
    \boldsymbol{\mathrm{b}}_t = [\boldsymbol{\theta}_1^{\mathrm{cmd}}, \boldsymbol{\theta}_2^{\mathrm{cmd}}, \boldsymbol{\theta}_3^{\mathrm{cmd}}, \boldsymbol{f}^{\mathrm{cmd}}, \boldsymbol{h}_z^{\mathrm{cmd}}, \boldsymbol{\phi}^{\mathrm{cmd}}, \boldsymbol{s}^{\mathrm{cmd}}, \boldsymbol{h}_z^{f, \mathrm{cmd}}]
\end{equation}
where $\boldsymbol{\theta}^{\mathrm{cmd}} = (\boldsymbol{\theta}_1^{\mathrm{cmd}}, \boldsymbol{\theta}_2^{\mathrm{cmd}}, \boldsymbol{\theta}_3^{\mathrm{cmd}})$ are the timing offsets between pairs of feet. Since our goal is to achieve pose tracking rather than diverse locomotion behaviors, we fix some gait parameters to speed up the convergence of training. In the following description, we specifically highlight the modified parts. 
 We set the timing offsets $\boldsymbol{\theta}^{\mathrm{cmd}}$ to $[0.5, 0, 0]$ to achieve a stable trotting gait. To enable the loco policy to recognize the rhythm of stepping, the clock time $\mathbf{t}_t$ is computed from the offset timings of each foot, with the mathematical definitions as follows: 
\begin{equation}
    \mathbf{t}_t = [\mathrm{sin}(2\pi t^{\mathrm{FR}}), \mathrm{sin}(2\pi t^{\mathrm{FL}}), \mathrm{sin}(2\pi t^{\mathrm{RR}}), \mathrm{sin}(2\pi t^{\mathrm{RL}})]
\end{equation}
\begin{equation}
    \begin{aligned}
        \relax [t^\mathrm{FR}, t^\mathrm{FL}, t^\mathrm{RR}, t^\mathrm{RL}] = [&t + \boldsymbol{\theta}_2^{\mathrm{cmd}} + \boldsymbol{\theta}_3^{\mathrm{cmd}}, t + \boldsymbol{\theta}_1^{\mathrm{cmd}} + \boldsymbol{\theta}_3^{\mathrm{cmd}}, \\
        &t + \boldsymbol{\theta}_1^{\mathrm{cmd}}, t + \boldsymbol{\theta}_2^{\mathrm{cmd}}]
    \end{aligned}
\end{equation}
where $t$ is a counter variable that advances from 0 to 1 during each gait cycle and $\mathrm{FR}, \mathrm{FL}, \mathrm{RR}, \mathrm{RL}$ are the four feet respectively. When the base velocity is zero, the jitter caused by marching on the spot will reduce the precision of manipulation. In this situation, we set clock time $\mathbf{t}_t $ to $[1, 1, 1, 1]$ to force all feet to maintain a stationary position. $\boldsymbol{f}^{\mathrm{cmd}}$ is the stepping frequency which is set to 3 Hz. $ \boldsymbol{h}_z^{\mathrm{cmd}}$ is the body height command which is not used. $\boldsymbol{s}^{\mathrm{cmd}}=(s^{\mathrm{cmd}}_x, s^{\mathrm{cmd}}_y)$ is the foot clearance which is set to $[0.45, 0.3]$. $\boldsymbol{h}_z^{f, \mathrm{cmd}}$ is the footswing height command which is set to 0.06 m.

During stage one, most components of the reward $r^{loco}$ utilized by the loco policy are identical to those described in \cite{margolis2023walk}, which can be referred to for more details. However, we remove the body height tracking component $r^{loco}_{\boldsymbol{h}_z^{\mathrm{cmd}}}$ to allow the body height to adjust adaptively according to the task. \blueline{The additional payload introduced by the arm increases the torque demands on the leg motors, which will result in motor overheating~\cite{ha2024umionlegs}. To mitigate this, we incorporate an energy reward $r^{loco}_{energy}$ for the quadruped robot to ensure the stability of instantaneous power~\cite{Fu2021MinimizingEC},} as formulated below:
\begin{equation}
    r^{loco}_{energy} = -0.00004 \cdot \sum_{i \in \text{leg joints}} \left| \tau_i \dot{q}_i \right|^2
\end{equation}
where $\tau_i$ and $\dot{q}_i$ are the torque and velocity of the $i^{th}$ leg joint.

\subsubsection{\textbf{Stage 2}} 
This stage aims to coordinate locomotion and manipulation to achieve whole-body large-range loco-manipulation. The arm policy is activated simultaneously with all the robotic arm joints. We adopt 6D target pose of end-effector as policy input. To eliminate the influence brought by body rotation \cite{fu2023deep}, we similarly use a posture-independent spherical coordinate to define the target end-effector pose $\chi_t$. The target position of end-effector is represented by radius $l$, latitude $p$, and longitude $y$. \blueline{The target orientation is represented using a 6-D vector~\cite{6d}.} To improve the accuracy of end-effector orientation tracking, we use euler angles $[roll, pitch, yaw]$ in Z-Y-X order
for sampling, which can intuitively exclude many illegal postures, and convert them to included angle along each axis of the coordinate. The mathematical form of sampling can be expressed as follows:
\begin{equation}
        R = R_{yaw} \cdot R_{pitch} \cdot R_{roll}
        = \begin{bmatrix}
        r_{11} & r_{12} & r_{13} \\
        r_{21} & r_{22} & r_{23} \\
        r_{31} & r_{32} & r_{33}
        \end{bmatrix}
\end{equation}

\begin{equation}
[\alpha, \beta, \gamma] = [\tan^{-1}(\frac{r_{21}}{r_{11}}), \tan^{-1}(\frac{r_{32}}{r_{22}}), \tan^{-1}(\frac{r_{13}}{r_{33}})]
\end{equation}
Here, $R$ is the composite rotation obtained by sequentially rotating around the z-axis, y-axis, and x-axis. $\gamma, \beta, \alpha$ represent the included angles with corresponding axes. To simultaneously minimize both position and orientation errors of the end-effector, the target pose tracking reward $r_{\chi_t}$ is constructed in exponential form:

\begin{equation}
    r_{\chi_t} = e^{-w \cdot \Delta(l, p, y)} \cdot e^{- \Delta (\alpha, \beta, \gamma)}
\end{equation}

\begin{equation}
 \left\{
 \begin{aligned}
    &\Delta(l, p, y) =  \sum_{u \in {(l, p, y)}}  k_{u} \cdot |u_t - u^{\mathrm{cmd}}|  \\
    &\Delta (\alpha, \beta, \gamma) =  \sum_{u \in { (\alpha, \beta, \gamma)}}   k_{u} \cdot |u_t - u^{\mathrm{cmd}}|
    \end{aligned}
 \right.
\end{equation}
where weight coefficient $w$ is used to balance the priority of the two components.  The function $\Delta(\cdot)$ is defined as the sum of absolute errors between each variable and its respective target value within the set, and $k_i$ represents the reciprocal of the sampling range for each variable, which is used to rescale the errors. The target pose tracking reward $r_{\chi_t}$ is utilized for both the loco and arm policies in stage 2.

\begin{table}[!ht]
\centering
\caption{Key Reward Terms, Equation, and \\Weights Used in Stage 1 and Stage 2}
\renewcommand{\arraystretch}{1.2}  
\setlength{\tabcolsep}{3pt}  
\begin{tabular}{>{\centering\arraybackslash}m{1.5cm} >{\centering\arraybackslash}m{4.2cm} >{\centering\arraybackslash}m{1cm} >{\centering\arraybackslash}m{1cm}} 
\toprule
\multirow{2}{*}{\raisebox{-0.5ex}{\textbf{Term}}} & \multirow{2}{*}{\raisebox{-0.5ex}{\textbf{Equation}}} & \multicolumn{2}{c}{\textbf{Weight}} \\
\cmidrule(lr){3-4}
& & \textbf{stage 1} & \textbf{stage 2} \\
\midrule
$r^{loco}_{\boldsymbol{s}^{\mathrm{cmd}}}$ & $\left(\boldsymbol{p}_{x, y, \text { foot }}^f-\boldsymbol{p}_{x, y, \text { foot }}^{f, \mathrm{cmd}}\left(\boldsymbol{s}^{\mathrm{cmd}}\right)\right)^2$ & -10.0 & 0.0\\

$r^{loco}_{hip}$ & $\sum\nolimits _{hip}|a_t|^2$ & 0.0 & -0.05\\

$r_{v^{\mathrm{cmd}}_{x,y}}$ & $\exp \left\{-\left|v_{x,y}-v_{x,y}^{\text {cmd }}\right|^2 / \sigma_{v_{x,y}}\right\}$ & 1.0 & 0.5\\

$r_{{\omega}^{\mathrm{cmd}}_{yaw}}$ & $\exp \left\{-\left({\omega}_{yaw} - {\omega}^{\mathrm{cmd}}_{yaw}\right)^2 / \sigma_{{\omega}_{yaw}}\right\}$ & 0.5 & 0.25\\

\midrule

$r^{arm}_{smooth}$ & $|a^{arm}_{t-1}-a^{arm}_{t}|^2$ & 0.0 & -0.1\\

$r^{arm}_{guide}$ & $|\boldsymbol{\phi}_t - a^{arm^G}_t |$ & 0.0 & -10.0\\

\bottomrule
\end{tabular}
\label{table:reward}
\end{table}

\subsubsection{\textbf{Reward Adjustment}} 
While the rewards in stage 1 drive the quadruped robot to achieve robust locomotion, the fixed foot placements are not conducive to the whole-body control required in stage 2, when the robot needs to adapt its body posture by altering foot placements. To address this, we replace the raibert heuristic footswing tracking reward $r^{loco}_{\boldsymbol{s}^{\mathrm{cmd}}}$ with the hip joint constraint reward $r^{loco}_{hip}$. Additionally, the velocity tracking rewards, $r_{v^{\mathrm{cmd}}_{x,y}}$ and  $r_{{\omega}^{\mathrm{cmd}}_{yaw}}$, are scaled to half of their original values to accommodate different tasks. The reward adjustments are detailed in Table \ref{table:reward}.
 To achieve more coordinated whole-body control, we introduce $r^{arm}_{smooth}$ and $r^{arm}_{guide}$ to promote smoother arm motion and enhance the control of the body orientation. \blueline{After the reward adjustments, the arm policy and locomotion policy will be synchronized to start stage 2 training.}

\section{Experiment Design}
\subsection{Robot System}
\begin{wrapfigure}{l}{0.25\textwidth}
    \setlength{\abovecaptionskip}{-4pt}
    \centering
    \includegraphics[width=0.25\textwidth]{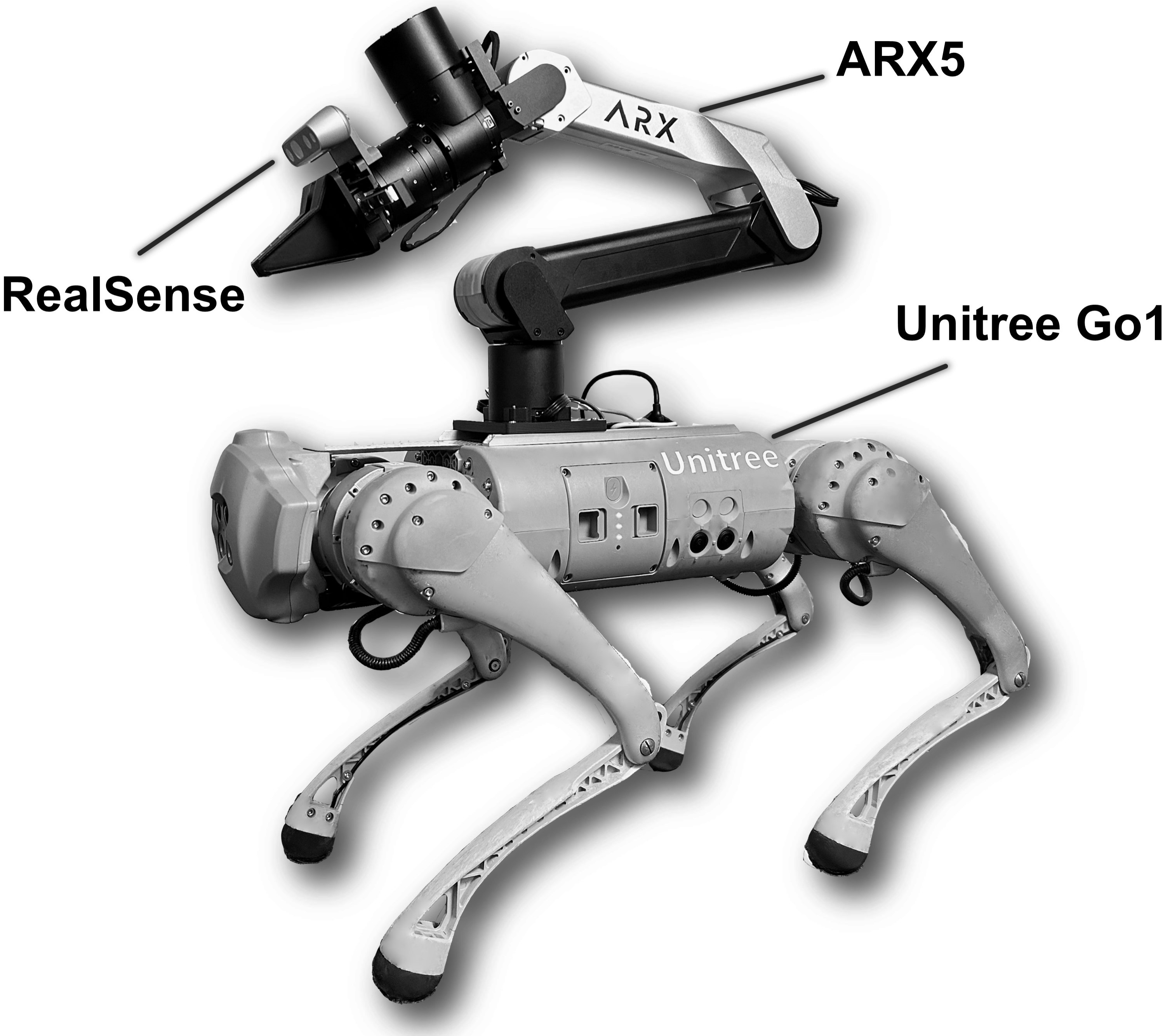}
    \caption{The robot system, consisting of a quadruped robot (Unitree Go1 Edu) with a mounted robotic arm (ARX5) and a camera (RealSense D435i).}
    \label{fig:system}
\end{wrapfigure}
The robot system used in our experiments consists of a 12-DoF legged robot Unitree Go1 Edu and a 6-DoF robotic arm ARX5 with a parallel gripper, as shown in Fig. \ref{fig:system}. The Go1 weighs 12 kg, and the ARX5, mounted on its back, weighs 3.35 kg. A RealSense D435i camera is positioned above the gripper to maintain a fixed relative pose. The ARX5 offers a rated load capacity of 1.5 kg and a maximum reach of 620 mm. The control frequency is set to 50 Hz for both training and deployment.

\subsection{Algorithm Comparison}
In order to examine the performance of our proposed method on whole-body control, 6D pose tracking, cross-embodiment, and loco-manipulation, we set up comparisons of the following algorithms:
\begin{itemize}
    \item \textbf{Floating Base (FB)+IK}: A floating base policy without whole-body control for the quadruped robot, combined with an IK solver~\cite{curobo_report23} for arm.
    \item \blueline{\textbf{Adaptive Floating Base (AFB)+ID}: An adaptive floating base policy coupled with an inverse dynamics based arm control strategy, corresponding to the built-in controllers of the Unitree Go1 and ARX5.}
    \item \textbf{Unified}: A unified policy with two output heads corresponding to the control of quadruped robot and arm. Training starts directly from stage 2.
    \item \textbf{Two-Stage}: The same policy settings as the Unified are used, but with a two-stage training approach.
    \item \textbf{Cooperated}: This setting uses the cooperative policy. Training starts directly from stage 2.
    \item \textbf{RoboDuet}: The same policy settings as the Cooperated are used, but with a two-stage training approach.
\end{itemize}
The first setting serves as a baseline to evaluate the solution stability of our method. \blueline{The second setting represents a more powerful real-world baseline, where the legged robot actively adjusts its posture in response to the interaction forces induced by the arm's motion.} The primary distinctions among the remaining four settings lie in whether they use a cooperative policy and a two-stage training approach, which are the two key components of our method. This comparison allows us to assess the effectiveness of each component.

\subsection{Training Details}
\blueline{We train 4096 parallel agents using the IsaacGym simulator~\cite{makoviychuk2021isaac} and the Proximal Policy Optimization (PPO) algorithm~\cite{schulman2017proximal}.} We train all algorithms for 50,000 iterations across 3 seeds, with the two-stage training comprising 10,000 iterations for stage 1 and 40,000 for stage 2, employing the asymmetric actor-critic framework used in~\cite{PintoAWZA18}. All neural networks are designed as the Multilayer Perceptron (MLPs) with ELU~\cite{Clevert2015FastAA} activations for hidden layers. \blueline{For the cooperative policy, the network architecture consists of three layers with 512, 256, and 128 neurons, respectively. In contrast to the cooperative policy,} the unified policy merges feature extractors into one, so we double the number of neurons in the hidden layers. All training is performed on NVIDIA RTX 4090 GPUs.

\subsection{Sample space}
The sample ranges utilized for both training and evaluation are detailed in Table \ref{table:range}. The evaluation sample space is slightly larger than the training one to demonstrate the generalization of algorithms, which already covers all 6D poses in the front hemisphere space. During the evaluation, we uniformly sample 200,000 target commands at random from the ranges to facilitate a comparative analysis of various algorithms.

\begin{table}[!ht]
\centering
\caption{Ranges of Commands Used in Training and Evaluation}
\renewcommand{\arraystretch}{1.3}
\label{table:range}
\begin{tabular}{ccc}
\toprule
\multirow{2}{*}{\raisebox{-0.5ex}{\textbf{Parameter}}} & \multicolumn{2}{c}{\textbf{Range}} \\

\cmidrule(lr){2-3}
                           & \textbf{Training}                & \textbf{Evaluation}          \\ 
\midrule
$v_x$ (m/s)                & {[}-1.00, 1.00{]}       & {[}-1.50, 1.50{]}   \\ 
$\omega_z$ (rad/s)         & {[}-0.60, 0.60{]}       & {[}-1.00, 1.00{]}   \\ 
$l$ (m)                    & {[}0.30, 0.70{]}        & {[}0.20, 0.80{]}    \\ 
$p$ (rad)                  & {[}-0.45$\pi$, 0.45$\pi${]}  & {[}-0.50$\pi$, 0.50$\pi${]} \\ 
$y$ (rad)                  & {[}-0.50$\pi$, 0.50$\pi${]}  & {[}-0.50$\pi$, 0.50$\pi${]} \\ 
$\alpha$ (rad)             & {[}-0.45$\pi$, 0.45$\pi${]}  & {[}-0.50$\pi$, 0.50$\pi${]} \\ 
$\beta$ (rad)              & {[}-0.33$\pi$, 0.33$\pi${]}  & {[}-0.50$\pi$, 0.50$\pi${]} \\ 
$\gamma$ (rad)             & {[}-0.42$\pi$, 0.42$\pi${]}  & {[}-0.50$\pi$, 0.50$\pi${]} \\ 
\bottomrule
\end{tabular}
\end{table}

\subsection{Metrics}
To quantify the performance of the algorithms, we define several metrics: (1) velocity tracking error, measured by the difference between the target commands and actual states for $v_x$, $v_y$, and $\omega_{yaw}$; (2) position tracking error, determined by the error for $l$, $p$, $y$, and the Euclidean distance $D$; (3) orientation tracking error, calculated based on the angles $\alpha$, $\beta$, $\gamma$, and the quaternion geodesic distance $\zeta$; (4) survival rate, assessed by randomly applying forces of 10 to 20 newtons to the base for 2 seconds and calculating the proportion of robots that maintain a base height above 0.26 m across all samples; (5) solvability, defined as the ratio of collision-free poses to the total number of samples; and (6) workspace, defined as the area of the convex hull formed by all target commands within the tracking error threshold, ensuring no self-collisions throughout the tracking process. A sample is considered successful if the tracking of the end-effector pose meets the thresholds of $D \leq 0.03$ m for distance and $\theta \leq \pi/36$ for orientation, where $\theta$ is the angle between the two vectors obtained by applying the target orientation and end-effector orientation to the same unit vector.  The maximum time allowed to reach the target command is 4 seconds, after which the average error over the following 2 seconds is computed.

\section{Results}

\subsection{Simulation Experiments}

\subsubsection{Whole-body Control} 
\begin{figure}[!ht]
    \setlength{\abovecaptionskip}{-4pt}
    \centering
    \includegraphics[width=0.48\textwidth]{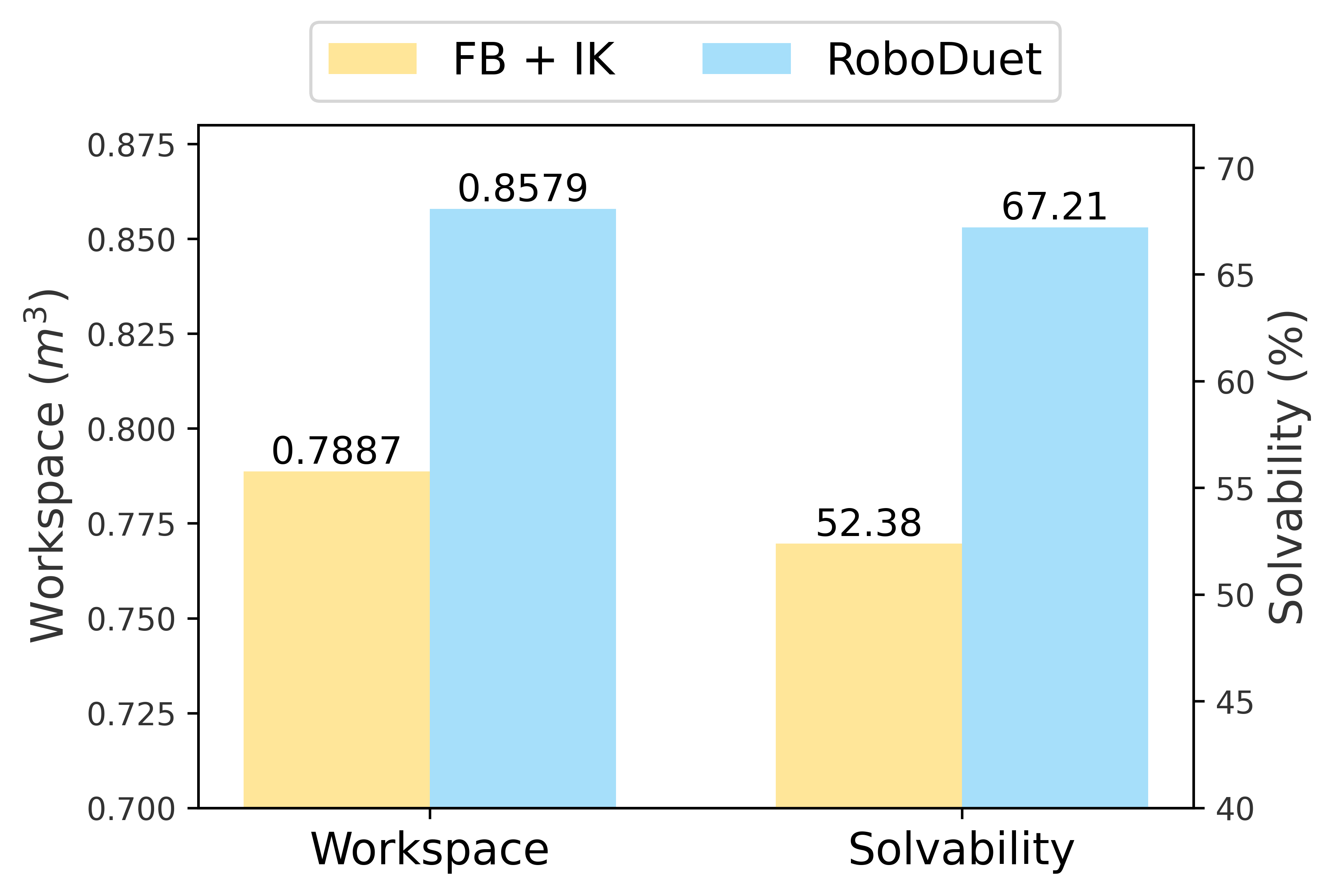}
    \caption{Comparison of Solvability and Workspace between FB+IK and RoboDuet.}
    \label{fig:sol}
\end{figure}

To validate the effectiveness of our proposed whole-body control training framework RoboDuet, we make a comparison with the FB+IK. To be fair, we evaluate the solvability and workspace under the same error threshold by applying the average Euclidean distance error and the quaternion geodesic distance error of \ourshort as the pose error thresholds for IK solver. For both, the initial joint positions of the arm are set to the same as default positions mentioned in Section \ref{subsubsec: stage-1}. According to Fig. \ref{fig:sol}, RoboDuet demonstrates a \blueline{14.83\%} improvement in solvability compared to the FB+IK configuration. Furthermore, the workspace comparison suggests that whole-body control can significantly enhance the operational capabilities of the robotic arm. It is worth reiterating that solvability is defined as reaching the target pose from the current pose without any collisions throughout the entire process. Additionally, many samples are inherently unreachable due to the limitations imposed by the structure of the robotic arm itself.

\begin{table*}[!ht]
\centering
\caption{Ablation Results \blueline{(scaled by $10^{-2}$)} of Different Methods in Still and Move Modes\\"Still" Means All Velocity Commands are Zero, While "Move" Involves Random Velocity Sampling During Evaluation.
}
\label{table:data}
\renewcommand{\arraystretch}{1.3}
\begin{tabular}{cc|cccc|cccc}
\toprule

\multicolumn{2}{c}{\multirow{2}{*}[-4pt]{\vspace{0.5em}\textbf{Metrics}}} & \multicolumn{4}{c}{\textbf{Still}} & \multicolumn{4}{c}{\textbf{Move}} \\
    \cmidrule(lr){3-6} \cmidrule(lr){7-10} 
    \multicolumn{2}{c}{}  
        & \blueline{\textbf{Unified}}    & \textbf{Two-Stage}   & \textbf{Cooperative} & \textbf{RoboDuet} & \blueline{\textbf{Unified}}    & \textbf{Two-Stage}   & \textbf{Cooperative} & \textbf{RoboDuet} \\ 
    \midrule
    \multirow{3}{*}{\begin{tabular}[c]{@{}c@{}}Velocity \\ Tracking $\downarrow$\end{tabular}} 
        & $v_x$ (m/s)        & \blueline{0.43$\pm$0.00} & \blueline{0.47$\pm$0.00} & \blueline{\textbf{0.30$\pm$0.00}} & \blueline{0.32$\pm$0.00} & \blueline{12.16$\pm$0.02} & \blueline{11.47$\pm$0.00} & \blueline{9.83$\pm$0.05} & \blueline{\textbf{9.7$\pm$0.00}} \\
        & $v_y$ (m/s)        & \blueline{0.45$\pm$0.00} & \blueline{0.44$\pm$0.00} & \blueline{0.36$\pm$0.00} & \blueline{\textbf{0.34$\pm$0.00}} & \blueline{17.10$\pm$0.00} & \blueline{16.84$\pm$0.02} & \blueline{15.91$\pm$0.00} & \blueline{\textbf{15.42$\pm$0.00}} \\ 
        & $\omega_z$ (rad/s) & \blueline{0.37$\pm$0.00} & \blueline{0.35$\pm$0.00} & \blueline{0.33$\pm$0.00} & \blueline{\textbf{0.32$\pm$0.00}} & \blueline{61.83$\pm$0.03} & \blueline{62.06$\pm$0.03} & \blueline{60.82$\pm$0.05} & \blueline{\textbf{60.59$\pm$0.02}} \\ 
    \midrule
    \multirow{4}{*}{\begin{tabular}[c]{@{}c@{}}Position \\ Tracking $\downarrow$\end{tabular}} 
        & $l$ (m)     & \blueline{4.25$\pm$0.04} & \blueline{4.13$\pm$0.00} & \blueline{\textbf{1.90$\pm$0.02}} & \blueline{1.97$\pm$0.00}        & \blueline{4.01$\pm$0.01}  & \blueline{4.25$\pm$0.01}  & \blueline{\textbf{1.88$\pm$0.00}} & \blueline{1.91$\pm$0.02} \\ 
        & $p$ (rad)   & \blueline{22.62$\pm$0.00} & \blueline{21.37$\pm$0.12} & \blueline{\textbf{17.42$\pm$0.03}} & \blueline{18.81$\pm$0.05}    & \blueline{21.93$\pm$0.11} & \blueline{22.45$\pm$0.07} & \blueline{\textbf{17.06$\pm$0.03}} & \blueline{18.09$\pm$0.01} \\ 
        & $y$ (rad)   & \blueline{12.01$\pm$0.01} & \blueline{11.52$\pm$0.04} & \blueline{7.94$\pm$0.00} & \blueline{\textbf{7.43$\pm$0.00}}      & \blueline{12.15$\pm$0.00} & \blueline{11.43$\pm$0.00} & \blueline{7.70$\pm$0.00} & \blueline{\textbf{7.22$\pm$0.00}} \\ 
        & $D$ (m)     & \blueline{14.47$\pm$0.01} & \blueline{13.84$\pm$0.01} & \blueline{\textbf{10.79$\pm$0.01}} & \blueline{11.08$\pm$0.00}    & \blueline{13.56$\pm$0.02} & \blueline{13.05$\pm$0.01} & \blueline{\textbf{10.58$\pm$0.00}} & \blueline{10.75$\pm$0.00} \\ 
    \midrule
    \multirow{4}{*}{\begin{tabular}[c]{@{}c@{}}Orientation \\ Tracking $\downarrow$\end{tabular}} 
        & $\alpha$ (rad) & \blueline{44.15$\pm$0.03} & \blueline{47.76$\pm$0.03} & \blueline{40.46$\pm$0.05} & \blueline{\textbf{39.17$\pm$0.02}}  & \blueline{42.50$\pm$0.08} & \blueline{47.63$\pm$0.07} & \blueline{43.22$\pm$0.04} & \blueline{\textbf{39.89$\pm$0.01}}\\ 
        & $\beta$ (rad)  & \blueline{60.41$\pm$0.13} & \blueline{62.86$\pm$0.08} & \blueline{48.89$\pm$0.03} & \blueline{\textbf{45.44$\pm$0.02}}  & \blueline{56.74$\pm$0.04} & \blueline{63.25$\pm$0.15} & \blueline{\textbf{45.29$\pm$0.01}} & \blueline{47.58$\pm$0.03} \\ 
        & $\gamma$ (rad) & \blueline{52.33$\pm$0.01} & \blueline{54.74$\pm$0.03} & \blueline{43.56$\pm$0.04} & \blueline{\textbf{39.38$\pm$0.04}}  & \blueline{52.36$\pm$0.03} & \blueline{54.38$\pm$0.05} & \blueline{42.74$\pm$0.02} & \blueline{\textbf{39.01$\pm$0.00}}\\ 
        & $\zeta$ (-)    & \blueline{51.15$\pm$0.01} & \blueline{53.46$\pm$0.02} & \blueline{48.18$\pm$0.00} & \blueline{\textbf{47.14$\pm$0.00}}  & \blueline{50.58$\pm$0.00} & \blueline{53.03$\pm$0.01} & \blueline{48.36$\pm$0.00} & \blueline{\textbf{47.53$\pm$0.00}}\\ 
    \midrule
    \multicolumn{2}{c|}{\textbf{Survival Rate (\%) $\uparrow$}}
                            & \blueline{87.35$\pm$0.78} & \blueline{95.66$\pm$0.06} & \blueline{94.49$\pm$0.03} & \blueline{\textbf{98.20$\pm$0.01}} & \blueline{93.03$\pm$1.34} & \blueline{99.09$\pm$0.01} & \blueline{98.56$\pm$0.02} & \blueline{\textbf{99.96$\pm$0.00}} \\ 
    \multicolumn{2}{c|}{\textbf{Workspace ($m^3$) $\uparrow$}}
                            & \blueline{51.36$\pm$0.46} & \blueline{60.10$\pm$0.05} & \blueline{84.08$\pm$0.06} & \blueline{\textbf{85.79$\pm$0.02}} & \blueline{78.96$\pm$0.04} & \blueline{81.30$\pm$0.02} & \blueline{86.71$\pm$0.01} & \blueline{\textbf{87.05$\pm$0.01}} \\ 

\bottomrule
\end{tabular}
\end{table*}

\subsubsection{Ablation} To validate the significance of the two-stage training and the cooperative mechanism, which are the key components of \ourshort, we conduct sufficient ablation experiments by comparing with Unified, Two-Stage and Cooperated. The results are shown in Table \ref{table:data}, which demonstrate that all configurations meet the requirements for stable standing. 
Although two-stage training does not significantly improve end-effector pose tracking, it demonstrates a notable improvement in resisting external perturbations. Specifically, Two-Stage improves survival rates by \blueline{8.31\% over the Unified in stationary standing}, attributed to the focus of stage 1 on optimizing locomotion. On the other hand, Cooperated and \ourshort outperform both the Unified and Two-Stage configurations across almost all metrics, indicating that the cooperative mechanism effectively decouples multi-task learning, enabling the two policies to collaborate while maintaining focus. \blueline{Cooperated achieves the best performance in position tracking}, largely due to its lack of the gait prior constraint from stage 1, allowing it to take a more aggressive approach in achieving higher tracking rewards which results a decrease on survival rates. 
In comparison, \ourshort maintains comparable end-effector tracking performance while achieving robust locomotion motion ability and offers a larger operational workspace during motion. In summary, by effectively integrating cooperative policy and two-stage training, RoboDuet significantly enhances control performance in both tracking accuracy and gait stability, underscoring the essential role of these components.

\begin{table}[!ht]
    \centering
    \caption{\blueline{Workspaces and survival rates of zero-shot policy transfer from Go1 to Go2 and A1.}}
    \label{table:embodiment}
    \begin{tabular}{ccccc}
    \toprule
    \multirow{2}{*}{\raisebox{-0.5ex}{\textbf{X+ARX}}} & \multicolumn{2}{c}{\textbf{Workspace ($m^3$)}} & \multicolumn{2}{c}{\textbf{Survival Rate ($\%$)}} \\

    \cmidrule(lr){2-3} \cmidrule(lr){4-5}
                           & \textbf{Still}          & \textbf{Move}          & \textbf{Still}              & \textbf{Move}             \\ 
    \midrule
       Go1  &    \blueline{\textbf{0.8579}}            &       \blueline{\textbf{0.8705}}           &         \blueline{\textbf{98.20}}           &    \blueline{\textbf{99.96}}              \\ 
    Go2  &    \blueline{0.8552}            &       \blueline{0.8216}           &         \blueline{96.32}           &    \blueline{93.21}             \\ 
    A1   &    \blueline{0.8416}            &       \blueline{0.8121}           &         \blueline{97.07}           &    \blueline{94.34}              \\ 
    \bottomrule
    \end{tabular}
\end{table}

\subsubsection{Zero-shot Transfer} To assess the zero-shot transfer capability of \ourshort, we introduce two additional quadruped robots, Unitree A1 and Unitree Go2, and mount the ARX5 on their backs. \blueline{The policy trained on the Go1+ARX system is directly transferred to these newly integrated platforms.} Subsequently, we evaluate their workspaces and survival rates in the presence of external disturbances, with the results presented in Table \ref{table:embodiment}. \blueline{Notably, under stationary standing conditions, Go2+ARX achieves a workspace of 0.8552, while A1+ARX attains a survival rate of 97.07\%, closely aligning with the performance of the original Go1+ARX system.} Due to fundamental differences among the various embodiments, only Go1+ARX5 demonstrates an increase in stability and workspace during motion through its custom-tailored cooperative policy. In contrast, the other two systems exhibit slight performance degradation. Nevertheless, they still achieve a survival rate of nearly \blueline{93\%} under external perturbations.


\begin{figure*}[!ht]
    \centering
    \vspace{0.3em}
    \includegraphics[width=0.98\textwidth]{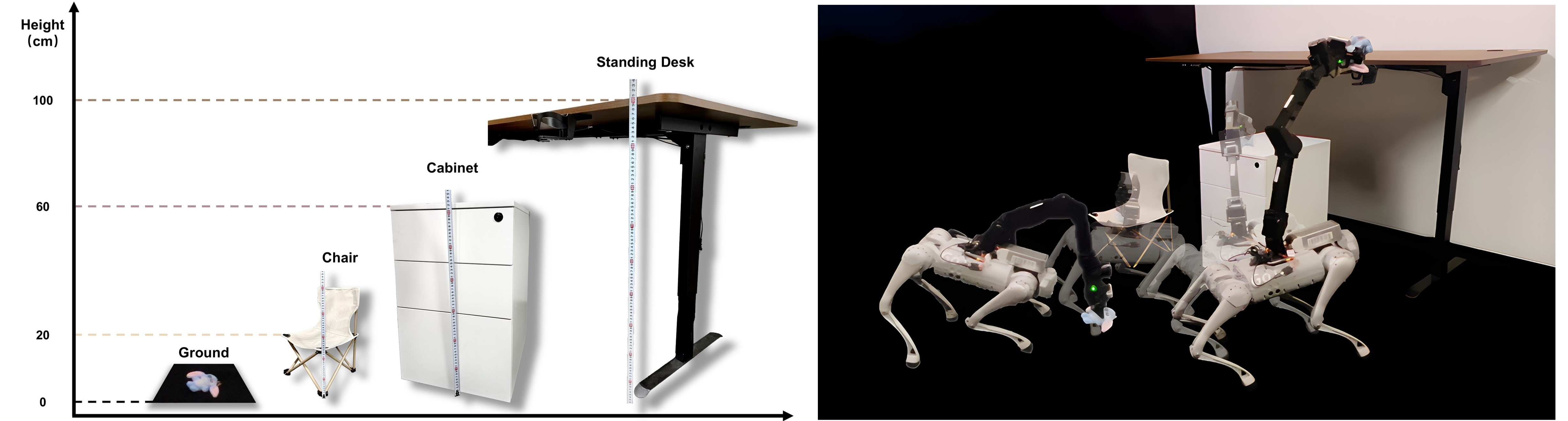}
    \caption{Transferring objects between different heights, including the ground (0 cm), a camping chair (20 cm), a cabinet (60 cm), and the cup holder of a standing desk (100 cm). The right part illustrates how RoboDuet utilizes whole-body control to adapt its posture for varying grasp poses during transfer tasks.}
    \label{fig:transferring}
\end{figure*}

\subsection{Real-world Experiments}
In terms of real-world experiments, we design three distinct types of tasks to evaluate the effectiveness of our policy in the real world and its proficiency in handling diverse loco-manipulation challenges. We directly deploy FB+IK, \blueline{AFB+ID,} and RobotDuet on a real-world robotic system for testing.

\begin{figure}[!ht]
    \centering
    \includegraphics[width=0.48\textwidth]{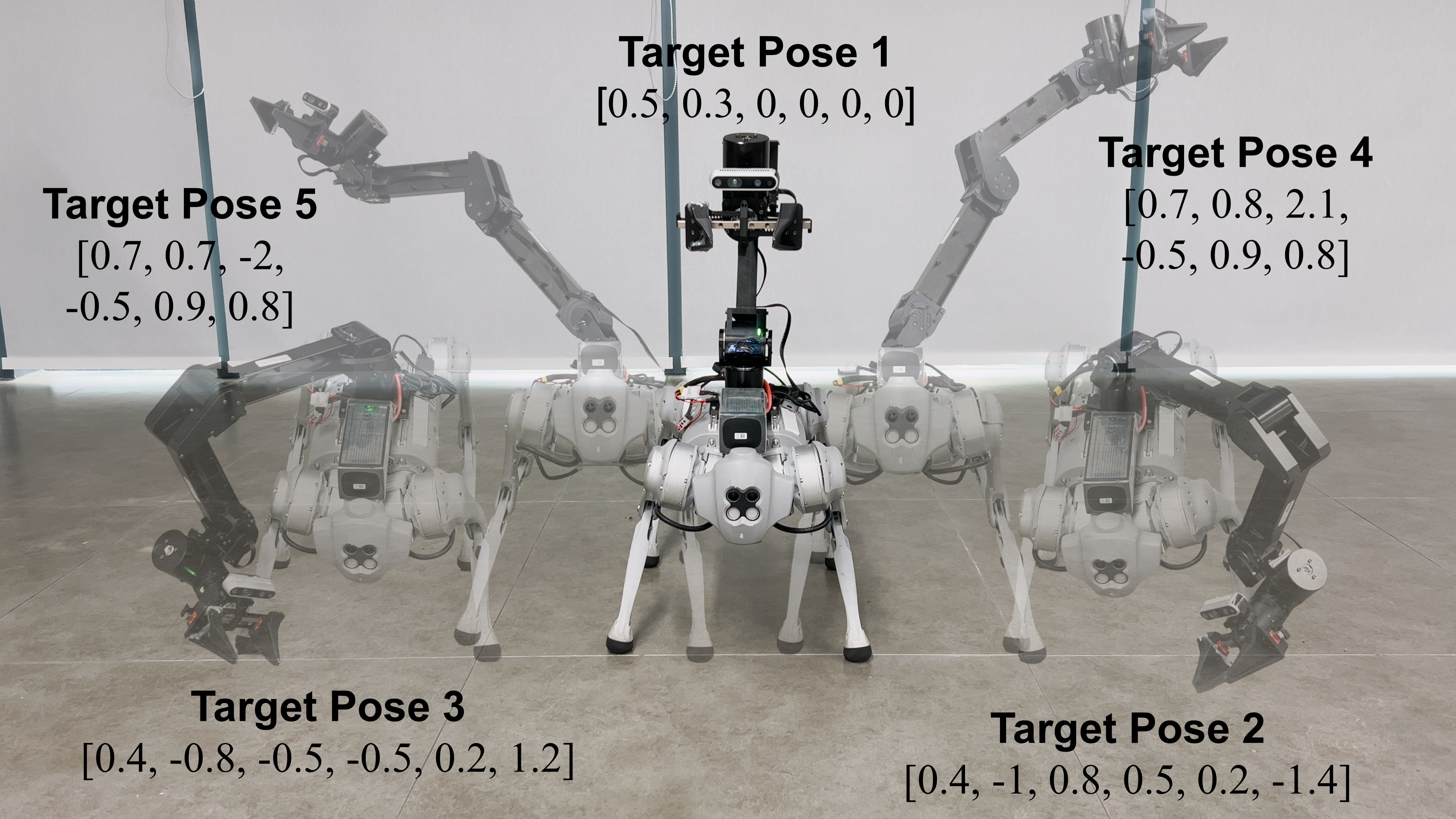}
    \caption{Tracking extreme 6D poses with RoboDuet. The target pose is represented as $[l, p, y, \text{roll}, \text{pitch}, \text{yaw}]$, with $l$ in meters and the others in radians. For certain unreachable targets, RoboDuet achieves stable approximations to optimal solutions.}
    \label{fig:extreme_pose}
\end{figure}

\subsubsection{Extreme 6D Pose Tracking}
To evaluate the robustness and generalization of end-effector pose tracking in our method, we select five target poses to compare the self-collision and solvability between  FB+IK and RoboDuet. The target poses are illustrated in Fig. \ref{fig:extreme_pose}, with four representing extreme 6D target configurations. Poses 2 and 3 are close to the ground and intersect with the robot's legs, while poses 4 and 5 are significantly elevated above the robot, beyond the training sampling range. Notably, pose 5 is located on the right side, nearing the robotic arm's maximum reach, and oriented towards the robot's head, representing an highly pathological configuration that is unsolvable. For each pose, we conduct eight repetitions, and the results are shown in Table \ref{tab:extreme_pose}. The IK system fails to resolve poses 2, 4, and 5. However, RoboDuet maintains stability across all poses and approached optimal approximate solutions for physically unreachable poses. Although pose 3 can be solved using IK, the floating base policy lacks whole-body control, leading to self-collision issues. Additionally, since the training does not include scenarios where the arm is in motion, tracking during poses 4 and 5 results in a higher risk of tipping, which we regard as self-collision. In contrast, due to the self-collision penalty integrated into the training process and its ability to maintain locomotion while the arm is in motion, RobotDuet experiences no collisions in nearly 95\% of trials, with the arm stopping effectively near the legs in poses 2 and 3. 

\begin{table}[!ht]
\centering
\caption{Self-collision and IK Failure Rate of \\
FB + IK vs. RoboDuet}
\label{tab:extreme_pose}
\renewcommand{\arraystretch}{1.3}
\begin{tabular}{ccc}
\toprule
\textbf{Method}           & \textbf{Self-collision Rate} & \textbf{IK Failure Rate} \\ 
\midrule
FB+IK & 50\%                & 60\%            \\ 
RoboDuet         & \textbf{5\%}                 & -               \\ 
\bottomrule
\end{tabular}
\end{table}

\subsubsection{Whole-body Control}
To further analyze the impact of whole-body control on loco-manipulation, we design a set of tasks involving transfering a doll from varying heights, as shown in Fig. \ref{fig:transferring}. \blueline{For each height, we calculate the average success rate from five repeated trials conducted by each of the three different operators.} A trial is considered successful only if the object is picked up and transferred to the next height without being dropped at any moment during this process. \blueline{The system's speed is controlled via the right VR handle, while the end-effector is operated with the left handle.}

\begin{table}[!ht]
    \centering
    \caption{Success Rates Across Different Heights for Various Control Methods in Object Transfer Tasks.}
    \renewcommand{\arraystretch}{1.3}
    \label{tab:real_wbc}
    \begin{tabular}{cccccc}
    \toprule
    \multirow{2}{*}{\raisebox{-0.5ex}{\textbf{Method}}} & \multicolumn{4}{c}{\textbf{Change in Height (cm)}} & \multirow{2}{*}{\raisebox{-0.5ex}{\textbf{Avg}}}\\
    \cmidrule(lr){2-5} 
    & \textbf{0 $\rightarrow$ 20} & \textbf{20 $\rightarrow$ 60} & \textbf{60 $\rightarrow$ 100} & \textbf{100 $\rightarrow$ 0} \\ 
    \midrule
    FB+IK & \blueline{3/15} & \blueline{7/15} & \blueline{0/15} & \blueline{0/15} & \blueline{10/60}\\
    [0.8ex]
    \blueline{AFB+ID} & \blueline{9/15} & \blueline{\textbf{13/15}} & \blueline{6/15} & \blueline{4/15} & \blueline{32/60}\\
    [0.8ex]
    RoboDuet & \blueline{\textbf{10/15}} & \blueline{12/15} & \blueline{\textbf{8/15}} & \blueline{\textbf{9/15}} & \blueline{\textbf{39/60}}\\ 
    \bottomrule
    \end{tabular}
\end{table}

The experimental results are shown in Table \ref{tab:real_wbc}. When grasping objects on the ground, it is crucial to avoid self-collisions. Lacking whole-body control capabilities, the FB+IK approach is prone to self-collisions and failure, particularly when picking up objects positioned in front of or below the head. \blueline{In contrast, AFB+ID and RoboDuet can adjust the body posture to minimize collisions during grasping, resulting in an almost 40\% increase in success rate. However, since AFB+ID can only adjust the rotation around the roll axis based on interaction forces, it still struggles to grasp objects positioned directly in front of the head. As the height increases, the need for body posture adjustments decreases, resulting in a improvement in success rates for all approaches.} Placing or retrieving the doll from the cup holder on a standing desk presented a particularly challenging task. The robotic arm was required to maintain a horizontal or top-down pose while operating near its maximum reach, when IK often fails to find a valid solution. \blueline{AFB+ID relies on its gravity compensation mechanism and inverse dynamics solution, achieving a placement success rate of 40\%. However, due to joint limit constraints, it can only successfully retrieve the doll when its ears or head are positioned near the outer edge of the cup holder.} RoboDuet demonstrats high solution stability and is able to reduce the vertical distance between the base and the standing desk by inclining the body, which successfully completed approximately \blueline{55\%} of the trials. \blueline{Compared to AFB+ID, RoboDuet achieves a 23\% improvement in success rate for more challenging loco-manipulation tasks.}

\subsubsection{Zero-shot Transfer}
\begin{figure}[!ht]
    \centering
    \setlength{\abovecaptionskip}{-4pt}
    \includegraphics[width=0.5\textwidth]{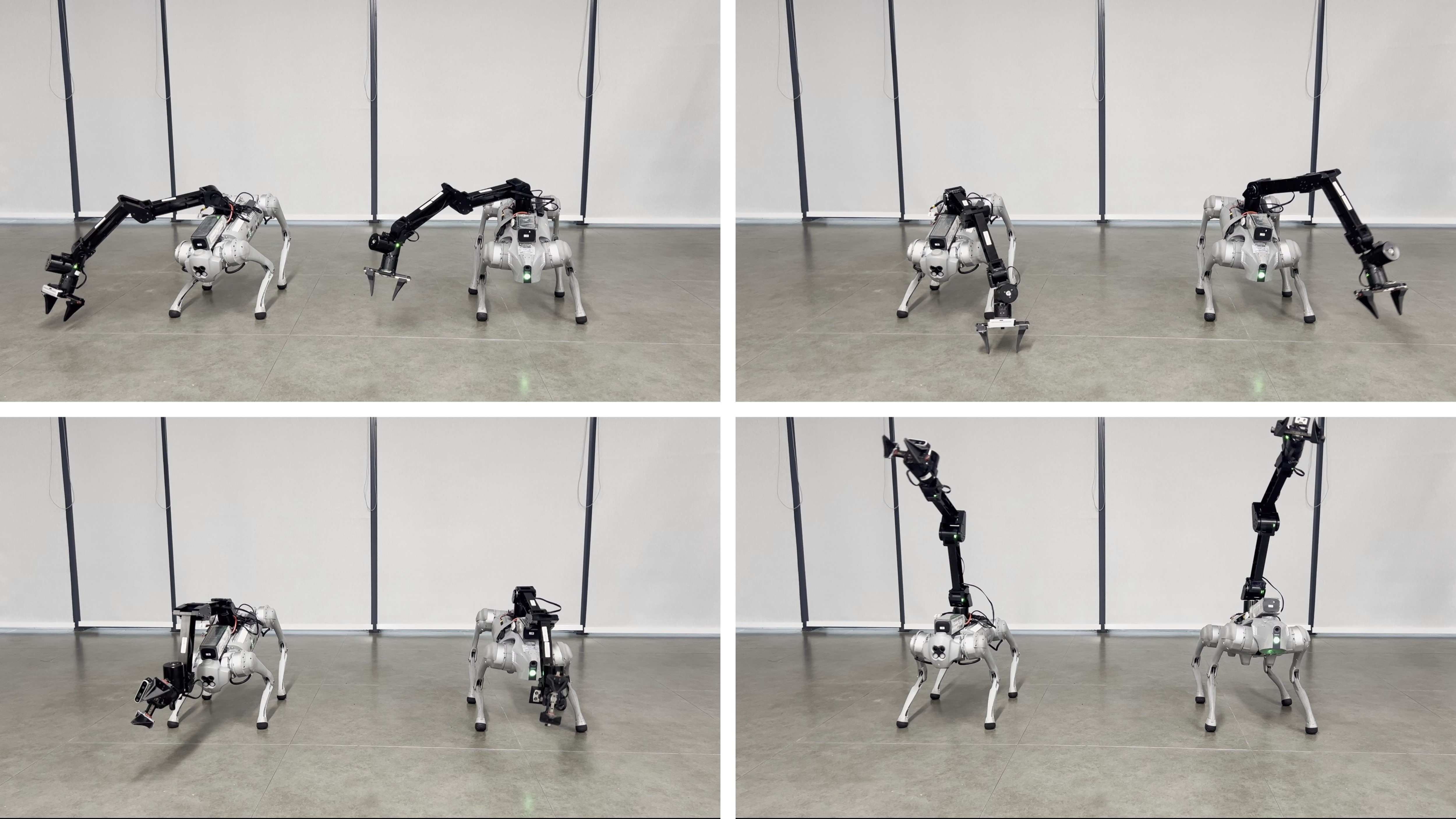}
    \caption{Zero-shot deployment of RoboDuet, trained on the Unitree Go1+ARX5 (left), onto the Unitree Go2+ARX5 (right). The bottom-left image illustrates consistent agile 6D pose tracking, while the others highlight coordinated whole-body control.}
    \label{fig:go2_go1}
\end{figure}

 To evaluate the generalizability of our method across different robot embodiments, we directly deploy the policy trained on the Unitree Go1+ARX5 to the Unitree Go2+ARX5. \blueline{The Unitree Go1 platform has a total mass of 18.35 kg, whereas the Unitree Go2 platform weighs 21.05 kg, representing a 14.7\% increase.} Despite the significant increase in mass, RoboDuet still demonstrated robust whole-body control across both platforms, as illustrated in Fig.~\ref{fig:go2_go1}. The figure depicts four key moments of the two systems tracking the same trajectory, demonstrating consistent 6D pose tracking and robust whole-body control capabilities across both configurations.

\section{Discussion and Limitations}
In this letter, we propose \ourshort, a whole-body legged loco-manipulation framework that integrates two collaborative policies for velocity tracking and 6D end-effector pose control, enabling agile and robust performance across diverse tasks. Leveraging a two-stage training approach, \ourshort effectively utilizes robust locomotion priors to enhance the system’s resistance to external disturbances. Additionally, our method supports \blueline{zero-shot transfer} deployment, allowing seamless hardware replacement without retraining, and it achieves real-world performance comparable to simulation.

While \ourshort focuses on low-level whole-body control, achieving fully autonomous task execution will require integration with high-level planners and advanced collision detection. \blueline{Moreover, our method can only handle terrains such as slopes and gravel while maintaining a fixed target pose. However, it has yet to achieve loco-manipulation on more complex terrains, such as stairs.} These are key areas for future enhancement.


\ifCLASSOPTIONcaptionsoff
  \newpage
\fi

\bibliographystyle{IEEEtran}
\balance
\bibliography{refer}

\end{document}